# A temporally abstracted Viterbi algorithm

**Shaunak Chatterjee**
Computer Science Division
University of California, Berkeley
Berkeley, CA 94720
shaunakc@cs.berkeley.edu

**Stuart Russell**
Computer Science Division
University of California, Berkeley
Berkeley, CA 94720
russell@cs.berkeley.edu

## Abstract

Hierarchical problem abstraction, when applicable, may offer exponential reductions in computational complexity. Previous work on coarse-to-fine dynamic programming (CFDP) has demonstrated this possibility using *state abstraction* to speed up the Viterbi algorithm. In this paper, we show how to apply *temporal abstraction* to the Viterbi problem. Our algorithm uses bounds derived from analysis of coarse timescales to prune large parts of the state trellis at finer timescales. We demonstrate improvements of several orders of magnitude over the standard Viterbi algorithm, as well as significant speedups over CFDP, for problems whose state variables evolve at widely differing rates.

## 1 Introduction

The Viterbi algorithm (Viterbi, 1967; Forney, 1973) finds the most likely sequence of hidden states, called the "Viterbi path," conditioned on a sequence of observations in a hidden Markov model (HMM). If the HMM has $N$ states and the sequence is of length $T$, there are $N^T$ possible state sequences, but, because it uses dynamic programming (DP), the Viterbi algorithm's time complexity is just $O(N^2T)$. It is one of the most important and basic algorithms in the entire field of information technology; its original application was in signal decoding but has since been used in numerous other applications including speech recognition (Rabiner, 1989), language parsing (Klein and Manning, 2003), and bioinformatics (Lytynoja and Milinkovitch, 2003).

Finding a most-likely state sequence in an HMM is isomorphic to finding a minimum cost path through a state–time *trellis graph* (see Figure 1) whose link cost is the negative log probability of the corresponding transition–observation pair in the HMM. Thus, the cost of finding an optimal path can be reduced further using an admissible (lower bound) heuristic and A* graph search.

Even with this improvement, the time and space cost can be prohibitive when $N$ and $T$ are very large; for example, with a state space defined by 30 Boolean variables, running Viterbi for a million time steps requires $10^{24}$ computations. One possible approach to handle such problems is to use a *state abstraction*: a mapping $\phi : S_0 \mapsto S_1$ from the original state space $S_0$ to a coarser state space $S_1$. For stochastic models (such as an HMM), the parameters of the model in $S_1$ are often chosen to be the maximum of the corresponding constituent parameters in $S_0$. Although these parameters do not define a valid probability measure, they serve as admissible heuristics for an A* search. The same idea can be applied to produce a a *hierarchy* of abstractions $S_0, S_1, \ldots, S_L$. Coarse-to-fine dynamic programming or CFDP (Raphael, 2001) begins with $S_L$ and iteratively finds the shortest path in the current (abstracted) version of the graph and refines along it until the current shortest path is completely refined. Several algorithms—e.g., hierarchical A* (Holte *et al.*, 1996) and HA*LD (Felzenszwalb and McAllester, 2007)—are able to refine only the necessary part of the hierarchy tree and compute heuristics only when needed.

The Viterbi algorithm devotes equal effort to every link in the state–time trellis. CFDP and its relatives can determine that an entire set of states need not be explored in detail, based on bounding the cost of paths through that set; *but they do so separately for each time step*. In this paper, we show how to use temporal abstraction to eliminate large sets of states from consideration *over large periods of time*.

To motivate our algorithm, consider the following problem. We observe Judy's daily tweets describing what she eats for lunch, and wish to infer the city in which she is staying on each day. The state space $S_0$ is the set of all cities in the world. The abstract space $S_1$ is the set of countries, and $S_2$ is the continents. (Figure 1 shows a small example.) The transition model suggests that on any given day Judy is unlikely to leave the city she is in, even less likely to leave the country she is in, and very unlikely indeed to travel to another continent. Thus, if Judy had Tandoori chicken on a Thursday but the rest of the week was all hamburgers, then it is most likely that she was in some American city *for the entire week*. However, if she had Tandoori chicken and/or biryani for an entire week, then it is quite possible that she is in India. Our algorithm, temporally abstracted Viterbi (henceforth TAV), facilitates *reasoning over a temporal interval* (like a week or month or longer) and *localized search within those intervals*. Neither of these is possible with Viterbi or state abstraction algorithms like CFDP. The computational savings of TAV on an instance of this problem can be seen in Figure 2.

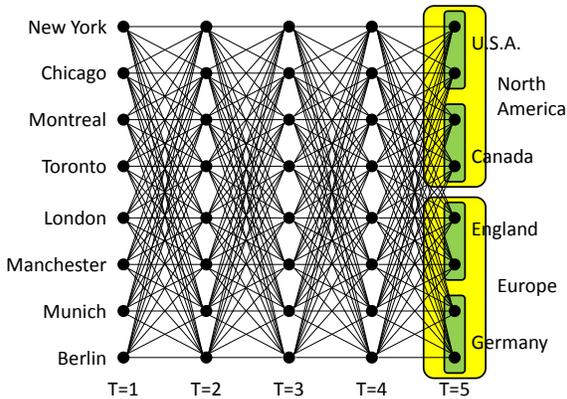

**Figure 1:** The state–time trellis for a small version of the tracking problem. The links have weights denoting probabilities of going from a city A to a city B in a day. The abstract state spaces $S_1$ (countries depicted in green) and $S_2$ (continents in yellow) are only shown for T=5 to maintain clarity. The observation links are also omitted for the same reason.

Temporal abstractions have been well-studied in the context of planning (Sutton *et al.*, 1999) and inference in dynamic Bayesian networks (Chatterjee and Russell, 2010). An excellent survey of temporal abstraction for dynamical systems can be found in (Pavliotis and Stuart, 2007). To the best of our knowledge, TAV is the first algorithm to use temporal abstraction for general shortest-path problems. TAV is guaranteed to find the Viterbi path, and does so (for certain problem instances) several orders of magnitude faster than the Viterbi algorithm and one to two orders of magnitude faster than CFDP.

The rest of the paper is organized as follows. Section 2 reviews the Viterbi algorithm and CFDP and introduces the notations and definitions used in the rest of the paper. Section 3 provides a detailed description of the main algorithm and establishes its correctness.

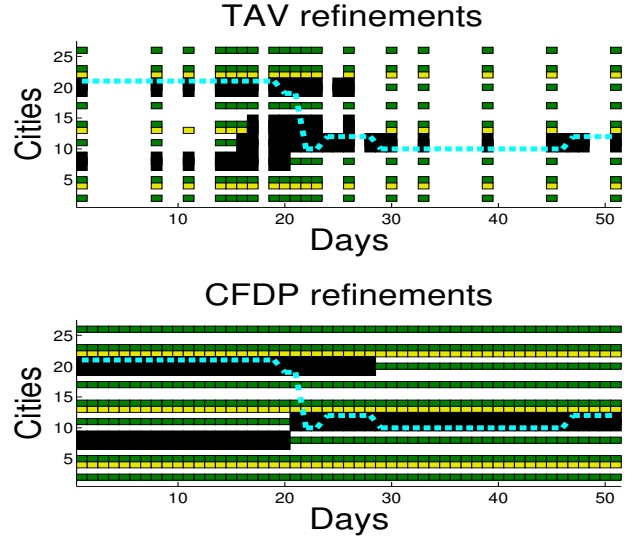

**Figure 2:** A comparison of the performance of CFDP and TAV on the city tracking problem with 27 cities, 9 countries and 3 continents over 50 days. The plots indicate portions of the state–time trellis each algorithm explored. Black, green and yellow squares denote the cities, countries and continents considered during search. The cyan dotted line is the optimal trajectory.

Section 4 discusses the computation of temporal abstraction heuristics. Section 5 presents some empirical results to demonstrate the benefits of TAV while section 6 provides some guidance on how to induce abstraction hierarchies.

## 2 Problem Formulation

Consider a hidden Markov model (HMM) whose latent Markovian state $X$ is in one of $N$ discrete states $\{1, 2, \ldots, N\}$. Let the actual state at time $t$ be denoted by $X_t$. The transition matrix $A = \{a_{ij} : i, j = 1, 2, \ldots, N\}$ defines the state transition probabilities where $a_{ij} = p(X_{t+1} = j \mid X_t = i)$. The Markov chain is assumed to be stationary, so $a_{ij}$ is independent of $t$. Let the discrete observation space be the set $\{1, 2, \ldots, M\}$. Let $Y_t$ be the observation symbol at time $t$. The observation matrix $B = \{b_{ik} : i = 1, 2, \ldots, N; k = 1, 2, \ldots, M\}$ defines the emission probabilities where $b_{ik} = p(Y_t = k \mid X_t = i)$. (We assume a discrete observation space in this paper, but our methods naturally extend to the continuous case.) The initial state distribution is given by $\Pi = \{\pi_1, \ldots, \pi_N\}$ where $\pi_i = p(X_0 = i)$.

The Viterbi path is the maximum likelihood sequence of latent states conditioned on the observation sequence. Following Rabiner (1989), we define

$$\delta_t(i) = \max_{X_{0:t-1}} p(X_{0:t-1}, X_t = i, Y_{1:t} \mid A, B, \Pi),$$

i.e., the likelihood score of the optimal (most likely) sequence of hidden states (ending in state $i$) and the

first $t$ observations. By induction on $t$, we have:

$$\delta_{t+1}(j) = [\max_i \delta_t(i) a_{ij}] b_{jY_{t+1}} .$$

The actual state sequence is retrieved by tracking the transitions that maximize the $\delta(.)$ scores for each $t$ and $j$. This is done via an array of back pointers $\psi_t(j)$. The complete procedure (Rabiner, 1989) is as follows:

1. Initialization
$$\begin{array}{rcl} \delta_1(i) & = & \pi_i\, b_{iY_1}, \quad 1 \leq i \leq N \\ \psi_1(i) & = & 0 \end{array}$$

2. Recursion

$$\begin{array}{rl} \delta_t(j) = & \max_{1 \leq i \leq N}[\delta_{t-1}(i) a_{ij}] b_{jY_t}, \quad 2 \leq t \leq T \\ \psi_t(j) = & \arg\max_{1 \leq i \leq N}[\delta_{t-1}(i) a_{ij}], \quad 2 \leq t \leq T \end{array}$$

3. Termination

$$\begin{array}{rcl} P^* & = & \max_{1 \leq i \leq N}[\delta_T(i)] \\ X_T^* & = & \arg\max_{1 \leq i \leq N}[\delta_T(i)] \end{array}$$

4. Path backtracking
$$X_t^* = \psi_{t+1}(X_{t+1}^*), \quad t = T-1, T-2, \ldots, 1$$

The time complexity of this algorithm is $O(N^2 T)$ and the space complexity is $O(N^2 + NT)$.

In coarse-to-fine (a.k.a. hierarchical) approaches, inference is performed in the coarser models to reduce the amount of computation needed in the finer models. Typically, a set of abstract state spaces $\mathcal{S} = \{S_0, S_1, \ldots, S_L\}$ and abstract models $\mathcal{M} = \{M_0, M_1, \ldots, M_L\}$ are defined where $S_0$ ($M_0$) is the original state space (model) and $S_L$ ($M_L$) is the coarsest abstract state space (model). Let the parameters of $M_l$ be denoted by the set $\{A^l, B^l, \Pi^l\}$.

A state in this hierarchy is denoted by $s_i^l$, where $l$ is the abstraction level and $i$ is its index within level $l$. $N_l$ denotes the number of states in level $l$. Let $\phi : S_l \mapsto S_{l+1}$ denote the mapping from any level $l$ to its immediate abstract level $l+1$. The parameters at level $l+1$ are defined by taking the maximum of the component parameters at level $l$. Thus, $A^{l+1} = \{a_{ij}^{l+1}\}$, where

$$a_{ij}^{l+1} = \max_{p,q} a_{pq}^l \quad s.t. \quad \phi(s_p^l) = s_i^{l+1}, \phi(s_q^l) = s_j^{l+1} .$$

$B^{l+1}$ and $\Pi^{l+1}$ are defined similarly in terms of $B^l$ and $\Pi^l$. Any transition/emission probability in an abstract model is a *tight* upper bound on the corresponding probability in its immediate refinement. Hence, the cost (negative log probability) of an abstract trajectory can serve as an admissible heuristic to guide search in a more refined state space.

CFDP works by starting with only the coarsest states $s_{1:N_L}^L$ at every time step from 1 to $T$. The states in $t$ and $t+1$ are connected by transition links whose values are given by $A^L$. $B^L$ and $\Pi^L$ define the other starting parameters. It then iterates between computing the optimal path in the current trellis graph and refining the states (and thereby the associated links) along the current optimal path. The algorithm terminates when the current optimal path contains only completely refined states (i.e., states in $S_0$). A graphical depiction of how CFDP works is shown in Figure 5. The TAV algorithm, which also has an iterative structure, is described in the next section. We will be reusing notation and definitions from this section throughout the rest of the paper.

## 3 Main algorithm

The distinguishing feature of TAV is its ability to reason with *temporally abstract links*. A link in the Viterbi algorithm and CFDP-like approaches describes the transition probability between states over a single time step. A temporally abstract link starting in state $s_1$ at time $t_1$ and ending in state $s_2$ at time $t_2 > t_1$ represents all trajectories having those end points and is denoted by a 4-tuple—$((s_1, t_1), (s_2, t_2))$. $Links((s, t))$ is the set of incoming links to state $s$ at time $t$. $Children(s^l) = \{s' : s' \in S^{l-1}, \phi(s') = s^l\}$ is the set of children of state $s^l$ in the abstraction hierarchy. We define three different kinds of temporally abstract links:

1. Direct links: $d(s, t_1, t_2)$ represents the set of trajectories that start at $(s, t_1)$ and end in $(s, t_2)$ and *stay within $s$* for the entire time interval $(t_1, t_2)$.

2. Cross links: $c((s_1, t_1), (s_2, t_2))$ represents the set of trajectories from $(s_1, t_1)$ to $(s_2, t_2)$, when $s_1 \neq s_2$.

3. Re–entry links: $r((s, t_1), (s, t_2))$ represents the set of trajectories that start at $(s, t_1)$ and end in $(s, t_2)$ but *move outside $s$ at least once* in the time interval $(t_1, t_2)$. $r((s_1, t_1), (s_1, t_2)) = \emptyset$ when $t_2 - t_1 \leq 1$.

The direct and cross links are denoted graphically by straight lines, whereas the re-entry links are represented by curved lines as shown in Figure 3. A generic link is denoted by the symbol $k$. The (heuristic) score of a temporally abstract link has to be an upper bound on the probability of all trajectories in the set of trajectories it represents. Computing admissible and monotone heuristics will be discussed in Section 4.

Our algorithm's computational savings over spatial abstraction schemes come from two avenues—first, fewer time points to consider using temporal abstraction; second, fewer states to reason about by considering *constrained trajectories* using direct links. Although the general flow of the algorithm is similar to CFDP, the refinement constructions are different. The algorithm descriptions provided omit details about observation matrix computations since they are standard. However, the issue is revisited in Section 4 to focus on some subtleties. The correctness of the algorithm depends only on the admissibility of the heuristics.

### 3.1 Refinement constructions

A refinement of a temporally abstract link replaces the original link with a set of refined links that represent a *partition*—a mutually exclusive and exhaustive decomposition—of the set of trajectories represented by the original link. The refinement allows us to reason about subsets of the original set of trajectories separately and thereby potentially narrow down on a single optimal trajectory. There are two different kinds of refinement constructions.

#### 3.1.1 Spatial refinement

When a direct link, $d(s^l, t_1, t_2)$, lies on the optimal path, the natural thing to do is to refine (partition) the set of trajectories it represents. The original direct link is *replaced* with all possible cross, direct and re-entry links between $Children(s^l)$ at $t_1$ and $t_2$. This is depicted graphically in Figure 3. A link is also refined spatially if its time span $(t_2 - t_1)$ is 1 time step since temporal refinement is not a possibility. The pseudocode for spatial refinement (see Algorithm 1) provides all the necessary details. It is trivial to show that the new links constitute a partition of the trajectories represented by the original link.

**Algorithm 1** Spatial Refinement$((p_1, t_1, p_2, t_2))$

$C \leftarrow Children(p_1); D \leftarrow Children(p_2)$
**if** $t_2 - t_1 > 1$ **then**
   $Links(p_1, t_2) \leftarrow Links(p_1, t_2) \setminus d(p_1, t_1, t_2)$
**else**
5:   $Links(p_2, t_2) \leftarrow Links(p_2, t_2) \setminus k((p_1, t_1), (p_2, t_2))$
**end if**
$usedStates(t_1) \leftarrow usedStates(t_1) \cup C$
$usedStates(t_2) \leftarrow usedStates(t_2) \cup D$
**for all** $s \in D$ **do**
10:   $Links(s, t_2) \leftarrow Links(s, t_2) \cup d(s, t_1, t_2)$
   **for all** $s' \in C$ **do**
     **if** $s = s'$ **then**
       $Links(s, t_2) \leftarrow Links(s, t_2) \cup r((s, t_1), (s, t_2))$
     **else**
15:       $Links(s, t_2) \leftarrow Links(s, t_2) \cup c((s', t_1), (s, t_2))$
     **end if**
   **end for**
**end for**

#### 3.1.2 Temporal refinement

When a cross link $c((s_1, t_1), (s_2, t_2))$ or a re-entry link $r((s_1, t_1), (s_1, t_2))$ is selected for refinement, we are faced with the task of refining a set of trajectories that *do not stay in the same state* for the abstraction interval. This is a case where temporal abstraction is not helping (not at the current resolution at least).

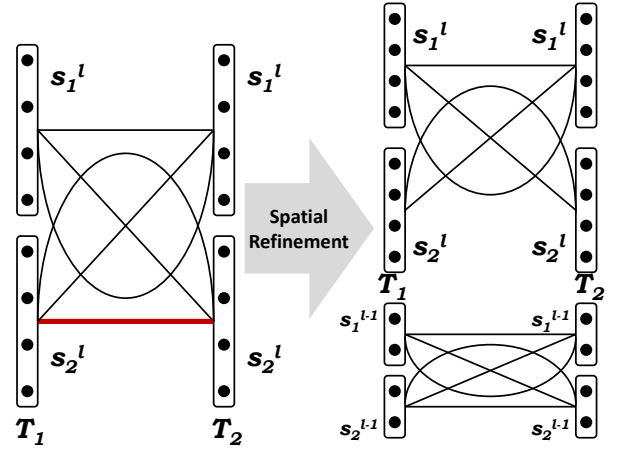

**Figure 3:** Spatial refinement: The optimal link, shown in bright red, is a direct link and is replaced with all possible links between its children.

An example of temporal refinement, which is only invoked when $t_2 - t_1 > 1$, is shown in Figure 4. It results in splitting the time interval $(t_1, t_2)$ into two sub-intervals that together span the original interval. Let us assume that we select the (rounded off) midpoint of the interval. When a link is temporally refined, we temporally split all cross, re-entry and direct links spanning the interval $(t_1, t_2)$ between states in $Children(\phi(s_1))$. We will show later that for any link longer than 1 time step, $\phi(s_1) = \phi(s_2)$. It should be noted that re-entry links are only added when the sub–interval length is longer than 1 time step. Also, if $t_2 - t_1 = 1$, then a cross link is *spatially refined* (analogous to CFDP).

One possibility is that some of the direct links for states in $Children(\phi(s_1))$ between $(t_1, t_2)$ were already spatially refined. In that case, we apply temporal refinement recursively to the spatially refined links of those direct links. *The choice of the splitting point does not affect the correctness of the algorithm as long as the split is replicated in the instantiated portion of the state space tree rooted at $\phi(s_1)$.* The pseudocode (shown in Algorithm 2) provides details of this procedure.

**Lemma 3.1** *The sets of trajectories represented by links before and after any spatial or temporal refinement are the same. Also, every trajectory is represented by exactly one temporally abstract path.*

**Lemma 3.2** *Any link created by TAV will always be between two states at the same level of abstraction. If the time span of the link is greater than 1 time step, then those two states will also have the same parent at all coarser levels of abstraction.*

**Proof** The original links are all between states of level $L$. Both refinement constructions add links only between states at the same abstraction level. This proves

**Algorithm 2** Temporal Refinement$((parent, t_1, t_2))$

    $nT \leftarrow \lceil(t_1 + t_2)/2\rceil$
    **if** $parent \in usedStates(nT)$ **then**
        **return**
    **end if**
5:  $usedTimes \leftarrow usedTimes \cup nT$
    $C \leftarrow Children(parent)$
    $usedStates(nT) \leftarrow usedStates(nT) \cup C$
    **for all** $s \in C$ **do**
        **if** $d(s, t_1, t_2) \notin Links(s, t_2)$ **then**
10:         $Temporal\_Refinement(s, t_1, t_2)$
            $Links(s, t_2) \leftarrow \emptyset$
            $Links(s, nT) \leftarrow \emptyset$
        **else**
            $Links(s, t_2) \leftarrow \{d(s, nT, t_2)\}$
15:         $Links(s, nT) \leftarrow \{d(s, t_1, nT)\}$
        **end if**
        **for all** $s' \in C$ **do**
            $Links(s, t_2) \leftarrow Links(s, t_2) \setminus k((s', t_1), (s, t_2))$
            $Links(s, t_2) \leftarrow Links(s, t_2) \cup k((s', nT), (s, t_2))$
20:         $Links(s, nT) \leftarrow Links(s, nT) \cup k((s', t_1), (s, nT))$
        **end for**
    **end for**

**Algorithm 3** BestPath$(Links, usedStates, usedTimes)$

    $curTime \leftarrow 1$
    **while** $curTime < T$ **do**
        $curTime \leftarrow nextUsedTime(curTime, usedTimes)$
        **for all** $s \in UsedStates(curTime)$ **do**
5:         $\delta_t(s) \leftarrow MaxOverLinks(Links((s, curTime)), \delta)$
            $\psi_t(s) \leftarrow ArgMaxOverLinks(Links((s, curTime)), \delta)$
        **end for**
        **for** $level = L - 1$ to $0$ **do**
            **for all** $s \in S^{level}$ && $s \in usedStates(curTime)$ **do**
10:             **if** $\delta_t(\phi(s)) > \delta_t(s)$ **then**
                $\delta_t(s) \leftarrow \delta_t(\phi(s))$
                $\psi_t(s) \leftarrow \psi_t(\phi(s))$
            **end if**
            **end for**
15:     **end for**
        **for** $level = 0$ to L-1 **do**
            **for all** $s \in S^{level}$ && $s \in usedStates(curTime)$ **do**
                **if** $\delta_t(\phi(s)) \leq \delta_t(s)$ **then**
                  $\delta_t(\phi(s)) \leftarrow \delta_t(s)$
20:                 $\psi_t(\phi(s)) \leftarrow \psi_t(s)$
                **end if**
            **end for**
        **end for**
    **end while**
25: $s^* \leftarrow \arg\max_{s \in usedStates(T)} \delta_T(s)$
    $curTime \leftarrow 1; \quad Path \leftarrow \emptyset$
    **while** $curTime > 1$ **do**
        $Path \leftarrow Path \cup \psi_{curTime}(s^*)$
        $(s^*, curTime) \leftarrow \psi_{curTime}(s^*)$
30: **end while**
    **return** $Path$

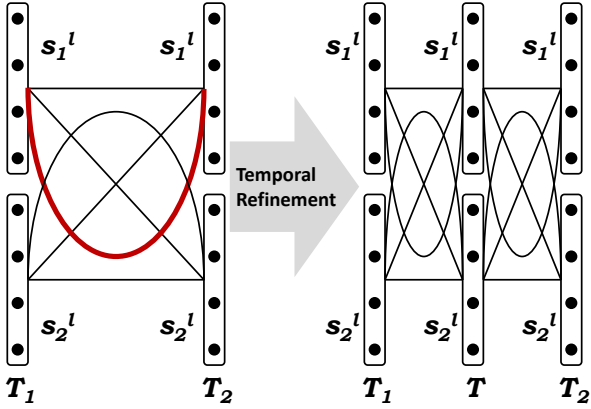

**Figure 4:** Temporal refinement: When refining a cross or re-entry link, refine all links between nodes that have the same parent as the nodes of the selected link.

the first statement. Moreover, upon initialization, there is no coarser level of abstraction—hence the second part of the statement is vacuously true. Temporal refinement always considers links between descendants of the *parent* node. Spatial refinement also adds links between $Children(s_l)$ of a state $s^l$ except when the time step is 1. This proves the second statement. □

### 3.2 Modified Viterbi algorithm

It is possible to have links to a state $s$ and to its abstraction $\phi(s)$ at the same time step $t$ (see Figure 5(b)). This was not possible in CFDP. Hence, we need a slightly modified scoring and backtracking scheme. $\delta_t(s)$ is the best score of a trajectory ending in state $s$ at time $t$ and $\psi_t(s)$ contains the temporally abstract link's information which connects $(s, t)$ to its predecessor. $usedTimes$ is a sorted list of time steps which have links to or from it. $usedStates(t)$ is the set of nodes at time $t$ which have incoming or outgoing links. The score computation algorithm moves forward in time like the normal Viterbi algorithm. The score computation (at each used time step $t$) is done in 3 phases. The pseudocode is given in Algorithm 3.

1. $\delta_t(s)$ is computed using the best of its incoming links, $Links(s, t)$ and $\psi_t(s)$ points to that link.

2. Starting at level $L - 1$ and going down to level 0, a state $s$ gets its parent's ($\phi(s)$) score and backpointer if $\phi(s)$ has a higher score.

3. Starting from level 0 and going up to level $L - 1$, a state $s$'s parent $\phi(s)$ gets its child's score and backpointer if $s$ has a higher score

**Theorem 3.3** *The score $\delta_t(s)$ computed by the BestPath procedure is a* strict upper bound *on all trajectories ending in state $s$ at time $t$ given the current abstracted version of the state-time trellis.*

**Proof** Any trajectory ending in a state $\hat{s}$, which is neither an ancestor nor a descendant of $s$, does not include any trajectory to $s$. Hence, *BestPath* computes an upper bound on the score of the best trajectory ending in state $s$ at time $t$.

For the bound to be strict, it is sufficient to show that each phase of *BestPath* only considers scores of such nodes where every incoming link includes at least one trajectory to $s$. The first phase accounts for all the incoming links to node $s$ itself. Let $\phi^*(s)$ denote an ancestor of $s$. Any incoming link (direct, cross or re–entry) to $\phi^*(s)$ includes at least one trajectory to $s$.

This necessitates taking the maximum over $\delta_t(\phi^*(s))$ (step 2). Finally, any trajectory ending in state $s'$, where $s'$ is a descendant of $s$, is by definition, a trajectory ending in $s$. Hence, the upper bound is strict. □

The ordering of the phases is important to perform the desired computation correctly and efficiently.

### 3.3 Complete algorithm

The algorithmic structure of TAV and CFDP are quite similar. The complete specification of TAV is presented in Algorithm 4. CFDP has a different initialization and refinement is node–based (TAV is link–based) which introduces links between states at different levels of abstraction. The two initializations are shown in Figure 5(a). CFDP's initial configuration has no temporally abstract links. The algorithm iterates between two stages: computing the optimal path in the current graph and refining links along the current optimal path. A few steps of execution of the two algorithms are shown on an example in Figure 5.

---

**Algorithm 4** TAV$(A, B, \Pi, \phi, Y_{1:T})$

$\delta_1(.) \leftarrow ScoreInitialization(\Pi)$
$usedStates(1) = usedStates(T) = S^L$
$usedTimes = \{1, T\}$
**for all** $s \in S^L$ **do**
5:  $Links(s, T) \leftarrow d(s, 1, T)$
    **for all** $s' \in S^L$ **do**
        $Links(s, T) \leftarrow Links(s, T) \cup k((s', 1), (s, T))$
    **end for**
**end for**
10: $ViterbiPathFound \leftarrow 0$
**while** $ViterbiPathFound = 0$ **do**
    $Path = BestPath(Links, usedStates, usedTimes)$
    $ViterbiPathFound = 1$
    **for all** $k \in Path$ **do**
15:     $((s_1, t_1), (s_2, t_2)) \leftarrow details(k)$
        **if** $level(s_1) > 1 \; || \; \neg isDirect(k)$ **then**
            $ViterbiPathFound = 0$
        **end if**
        **if** $isDirect(k) \; || \; t_2 - t_1 = 1$ **then**
20:         $Spatial\ Refinement(k)$
        **else**
            $Temporal\ Refinement(k)$
        **end if**
    **end for**
25: **end while**

---

The correctness of the algorithm follows from the optimality of A* search and Lemma 3.1 and Theorem 3.3.

## 4 Heuristics for temporal abstraction

In hierarchical state abstraction schemes, computing heuristics involves taking the maximum of a set of single time step transition probabilities. As mentioned in Section 2, this can be done by hierarchically constructing $A_l, B_l$ and $\Pi_l$. For temporal abstractions however, there are more design choices to be made when it comes to computing heuristic scores of links. There is a more significant tradeoff between cost of computation and quality of heuristic.

The heuristic score of a direct link is very easy to compute. We do not have to select between possible state transitions. Thus, the heuristic for a link spanning the interval $(t_1, t_2)$ can be done in $O(t_2 - t_1)$, since we still need to account for all the observations in that interval. If the score is cached, the score for any subinterval is computable in $O(1)$ time.

Cross links and re–entry links require further consideration. A somewhat expensive option is to compute the Viterbi path in the restricted scope. As Lemma 3.2 shows, a cross or a re–entry link represents trajectories that can switch between sibling states (the ones which map to the same parent via $\phi$). In an abstraction hierarchy, if the cardinality of $Children(s)$ for any state $s$ is restricted to some constant $C$, then computing this heuristic will require $O(C^2(t_2 - t_1))$ time.

A computationally cheaper but relatively loose heuristic is the following:

$$h((s_i, t_1), (s_j, t_2)) = \max_k \hat{A}_{ik} \max_{p,q} \hat{A}_{pq}^{t_2-t_1-2} \max_k \hat{A}_{kj}$$
$$\prod_t \max_k \hat{B}_{kY_t}$$

$\hat{A}$ and $\hat{B}$ represent the transition matrices for the set $Children(\phi(s_i))$. This heuristic chooses the best possible transition at every time step other than the two end points and also the best possible observation probability. Its computational complexity is $O(C^2 + (t_2 - t_1))$. Caching values help in both cases. The Viterbi heuristic, being tighter, leads to fewer iterations but needs more computation time. We will compare the two heuristics in our experiments.

## 5 Experiments

The simulations we performed were aimed at showing the benefit of TAV over Viterbi and CFDP. The benefits are magnified in systems where variables evolve at widely varying timescales. The timescale of a random variable is the expected number of time steps in which it changes state. A person's continental location would have a very large timescale, whereas his zip code location would have a much smaller timescale.

A natural way to generate transition matrices with timescale separation is to use a dynamic Bayesian network (DBN). We consider a DBN with $n$ variables, each of which has a cardinality of $k$. Hence, the state space size $N$ is $k^n$. We used fully connected DBNs in our simulations. Our observation matrix was multimodal (hence somewhat informative). A DBN with a parameter $\epsilon$ means that the timescales of successive variables have a ratio of $\epsilon$. The fastest variable's timescale is $1/\epsilon$ and the slowest variable's $(1/\epsilon)^k$.

The state space hierarchy at the most abstract level arises from the branching of the slowest variable. In each subtree, we branch on the next slowest variable. In the experiments in this section we assume that the abstraction hierarchy is given to us.

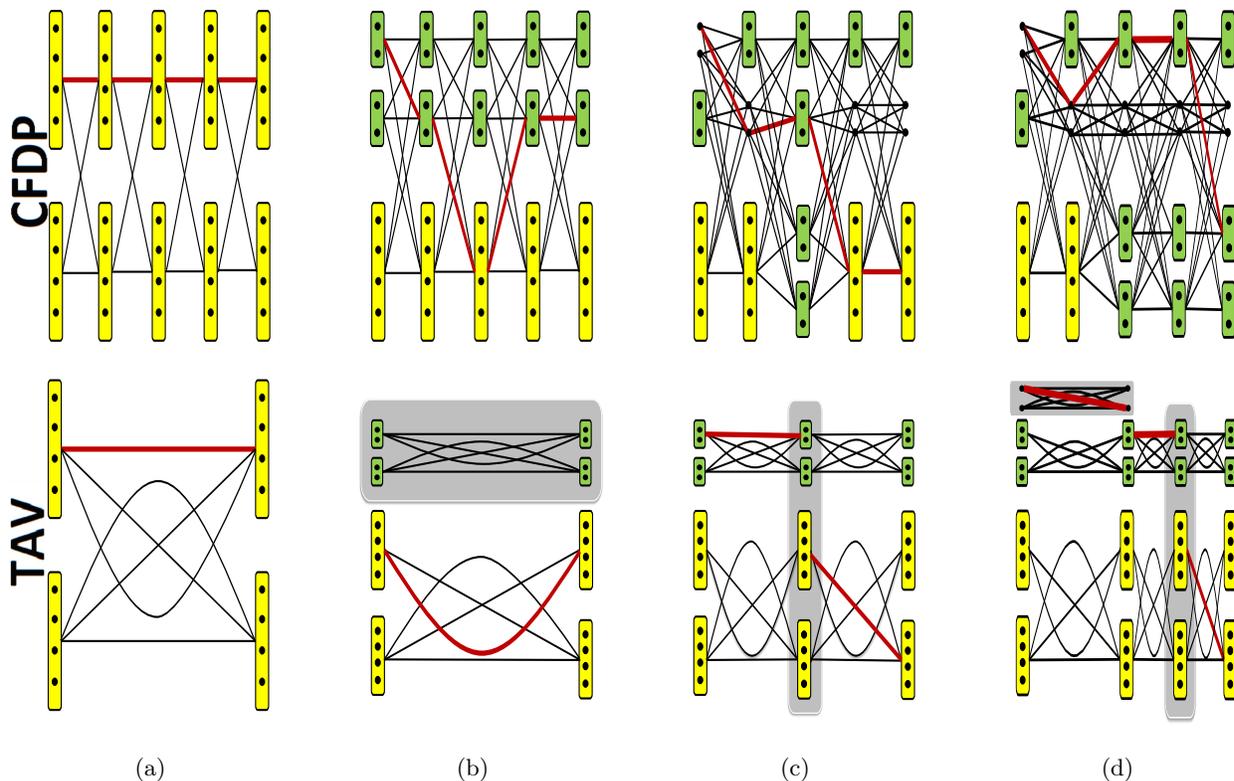

**Figure 5:** Sample run: TAV: (a) Initialization. The optimal path is a direct link—hence spatial refinement. The new additions are shadowed. (b) A re-entry link is optimal—hence temporal refinement. Since one direct link among siblings was already refined in Step 1, we also temporally refine the spatially refined component. (c) The optimal path has links at different levels of abstraction. Such scenarios necessitate the $BestPath$ procedure. (d) More recursive temporal refinement is performed. Note the difference in the numbers of links in the two graphs after 3 iterations.

### 5.1 Varying T, N and $\epsilon$

To study the effect of increasing $T$ on the computation time, we generated 2 sequences of length 100000 with $\epsilon = .1$ (case 1) and .05 (case 2). $N$ was 256 and the abstraction hierarchy was a binary tree. For each sequence, we found the Viterbi path for the first $T$ timesteps using TAV, CFDP and the Viterbi algorithm. The results are shown in Figure 6(a). TAV's computational complexity is marginally super-linear. This is because TAV might need to search in the interval $[0, T]$ even if it had found the Viterbi sequence in that interval as new observations (after time $T$) come in. For the $\epsilon$ values used, TAV is more than 2 orders of magnitude faster than Viterbi and 1 order of magnitude faster than CFDP.

It is intuitive that TAV will benefit more from a smaller $\epsilon$ (i.e. a larger timescale). As Figure 6(a) shows, CFDP also benefits from the timescale artifact. However, TAV's gains are larger and also grow more quickly with diminishing $\epsilon$. This set of experiments had $N = 256$ and $T = 10000$.

Finally, to check the effect of the state space size, we chose $\epsilon = .5$ (case 1) and .25 (case 2). We chose fast timescales to show the limitations of TAV (having already demonstrated its benefits for small $\epsilon$). As Figure 6(a) shows, TAV is 3x to 10x faster than CFDP for $N < 1024$. However, CFDP is about 6x faster than TAV for $N = 2048$. In this case, TAV performs poorly because of its initialization as a single interval. For fast timescales, the first few refinements in this setting are invariably temporal and these refinements can be computationally very expensive.

### 5.2 *A priori* temporal refinement

If $T$ is comparable to the timescale of the slowest variable, then one or more temporal refinements at the very outset of TAV is very likely. Performing these refinements *a priori* will be beneficial if TAV actually had to perform those refinements. The benefit would be proportional to the cost of deciding whether to refine or not. This decision cost increases with $T$ and $N$ (but does not depend on $\epsilon$).

Figure 6(b) shows the computation time for cases where we initialized TAV with 20 equal segments. The *a priori* refinement time was also included in the "Pre-Segmentation" time. Speedups of 2x to 6x were ob-tained for varying values of $N$ and $T$. When only $\epsilon$ was varied, the benefit was approximately constant (between 3 to 6 seconds of computation time). This resulted in effective speedups only for the smaller values of $\epsilon$, which had small computation times.

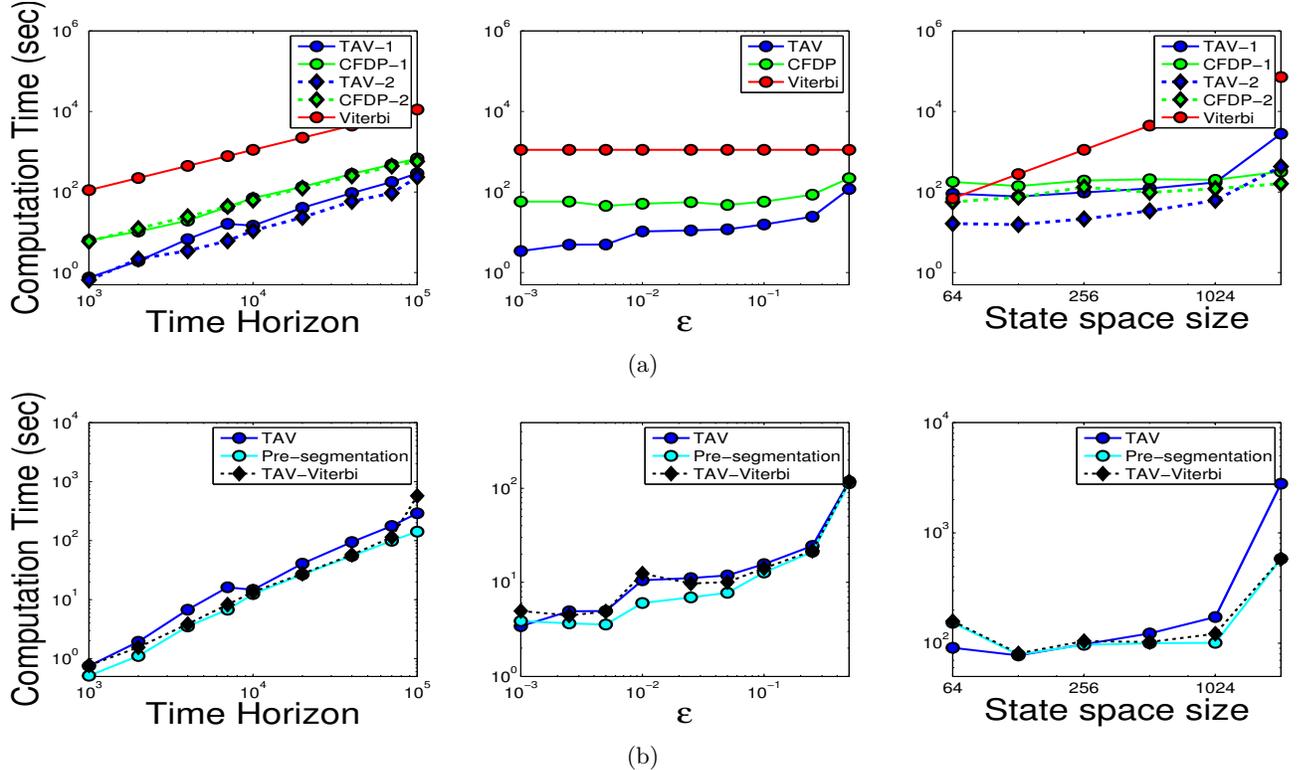

**Figure 6:** Simulation results: (a) The computation time of Viterbi, CFDP and TAV with varying $T$ (left), $\epsilon$ (middle) and $N$ (right). (b) The computation time of TAV and its two extensions—pre-segmentation and using the Viterbi heuristic—with varying $T$ (left), $\epsilon$ (middle) and $N$ (right).

### 5.3 Impact of heuristics

As discussed in Section 4, there is a trade-off in computing heuristics between accuracy and computation time. Figure 6(b) compares the effect of using the Viterbi heuristic instead of the cheap heuristic described previously. With increasing $T$, there was a small improvement in computation time, although the speedup was never greater than 2x. The two computation times were virtually the same with increasing $\epsilon$. For large state spaces, the Viterbi heuristic produced more than 5x speedup (which made it comparable to CFDP).

The main reason for the lack of improvement (in computation time) is the randomness of the data generation process. The Viterbi heuristic can significantly outperform the cheap heuristic only if the most likely state sequences according to the transition model receive very poor support from the observations. In that case, the cheap heuristic will provide very inaccurate bounds and mislead the search. In randomly generated models however, the two heuristics demonstrate comparable performance.

## 6 Hierarchy induction

Till this point, we have assumed that the abstraction hierarchy will be an input to the algorithm. However, in many cases, we might have to construct our own hierarchy. Spectral clustering (Ng *et al.*, 2001) is one technique which we have used in our experiments to successfully induce hierarchies. If the underlying structure is a DBN of binary variables with timescale separation between each variable (as discussed in Section 5), there will be a gap in the eigenvalue spectrum after the two largest values. The first two eigenvectors will be analogous to indicator functions for branching on the slowest variable. We can then apply the method recursively within each cluster. The main drawback is the $O(N^3)$ computational cost. It can be argued that this is an offline and one-time cost—nonetheless it is quite expensive. It should be noted that spatial hierarchy induction is also a hard problem.

Let $2^8$ denote a 8-variable DBN where each variable is binary. The slowest variables are placed at the left—hence $4^2 2^4$ denotes a DBN whose two slowest variables are 4-valued. As we change the underlying model from $2^8$ to $4^4$ to $16^2$, is there a particular abstraction hierarchy which performs well for all the models?

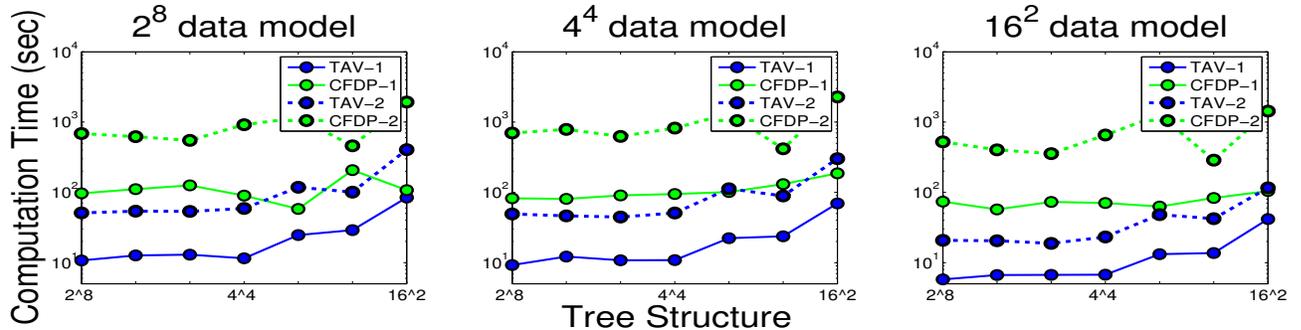

**Figure 7:** Effect of abstraction hierarchy: For different underlying models ($2^8$, $4^4$ and $16^2$), deep hierarchies outperform shallow hierarchies. Cases 1 and 2 have $\epsilon = 0.1$ and .05 respectively

For the experiments, we generated 3 different data sets using the following DBNs—$2^8$ (left), $4^4$ (middle) and $16^2$ (right)—with $N = 256$. On each data set, we used the following abstraction hierarchies ($2^8$; $2^4 4^2$; $4^2 2^4$; $4^4$; $4^1 8^2$; $8^2 4^1$; $16^2$). The results in Figure 7 show the computation time for TAV and CFDP using different abstraction hierarchies (deepest on the left to shallowest on the right) for two different values of $\epsilon$. Both TAV and CFDP *perform better with deeper hierarchies*, although the improvement is much more pronounced for TAV. The trend across all 3 underlying data models indicates that we could always induce a deep hierarchy. The benefit of lightweight local searches in a deep hierarchy seems to outweigh the cost of the necessary additional refinements.

## 7 Conclusion

We have presented a temporally abstracted Viterbi algorithm, that can reason about sets of trajectories and uses A* search to provably reach the correct solution. Direct links provide a way to reason about trajectories within a set of states—something that previous DP algorithms did not do. For systems with widely varying timescales, TAV can outperform CFDP handsomely. Our experiments confirm the intuition—the greater the timescale separation, the more the computational benefit.

Another smart feature of our algorithm is that it can exploit multiple timescales present in a system by adaptive spatial and temporal refinements. TAV's limitations arise when the system has frequent state transitions and in such cases, it is better to fall back on the conventional Viterbi algorithm (CFDP is often slower as well in such cases). It might be possible to design an algorithm that uses temporal abstraction and can also switch to conventional Viterbi when the heuristic scores of direct links are low.

### Acknowledgements

We would like to acknowledge NSF for their support (grant no. IIS-0904672). We would also like to thank Jason Wolfe, Aastha Jain and the anonymous reviewers for their comments and suggestions.